\documentclass[letterpaper, 10 pt, conference]{ieeeconf}  

\IEEEoverridecommandlockouts                              

\usepackage{graphics} 
\usepackage{times} 
\usepackage{amsmath} 
\usepackage{amssymb}  
\usepackage{amsfonts}
\usepackage{subfigure}
\usepackage{hyperref}
\usepackage{graphicx}
\usepackage{float}
\usepackage{caption}
\usepackage{multirow}
\usepackage[ruled,linesnumbered]{algorithm2e}
\usepackage{color}
\hypersetup{hidelinks}
\usepackage{booktabs}

\def\ralhigh#1{\textcolor{black}{#1}}

\title{\LARGE \bf
DexTOG: Learning Task-Oriented Dexterous Grasp with Language Condition} 

\author{Jieyi Zhang$^{1}$, Wenqiang Xu$^{1}$, Zhenjun Yu$^{1}$, Pengfei Xie$^{2}$, Tutian Tang$^{1}$ and Cewu Lu$^{1}$
\thanks{$^{1}${\tt\small \{yi\_eagle, vinjohn, jeffson-yu, tttang, lucewu\}@sjtu.edu.cn}. Jieyi Zhang, Wenqiang Xu, Zhenjun Yu, Tutian Tang are with the School of Electronic Information and Electrical Engineering, Shanghai Jiao Tong University, Shanghai, China. Cewu Lu is the corresponding author, a member of Qing Yuan Research Institute and MoE Key Lab of Artificial Intelligence, AI Institute, Shanghai Jiao Tong University, Shanghai, China.}
\thanks{$^{2}${\tt\small xiepf2002@gmail.com}. Pengfei Xie is with Southeast University.
}}

\begin{document}

\maketitle
\thispagestyle{empty}
\pagestyle{empty}

\begin{abstract}
This study introduces a novel language-guided diffusion-based learning framework, \ralhigh{DexTOG}, aimed at advancing the field of task-oriented grasping (TOG) with dexterous hands. Unlike existing methods that mainly focus on 2-finger grippers, this research addresses the complexities of dexterous manipulation, where the system must identify non-unique optimal grasp poses under specific task constraints, cater to multiple valid grasps, and search in a high degree-of-freedom configuration space in grasp planning. \ralhigh{The proposed DexTOG includes a diffusion-based grasp pose generation model, DexDiffu, and a data engine to support the DexDiffu. By leveraging DexTOG, we also proposed a new dataset,} DexTOG-80K, which was developed using a shadow robot hand to perform various tasks on 80 objects from 5 categories, showcasing the dexterity and multi-tasking capabilities of the robotic hand. This research not only presents a significant leap in dexterous TOG but also provides a comprehensive dataset and simulation validation, setting a new benchmark in robotic manipulation research. You can find more details on the website: \url{https://sites.google.com/view/dextog}.

\end{abstract}

\section{Introduction}
\label{sec:intro}
\begin{figure}[t!]
    \centering
    \includegraphics[width=0.8\linewidth]{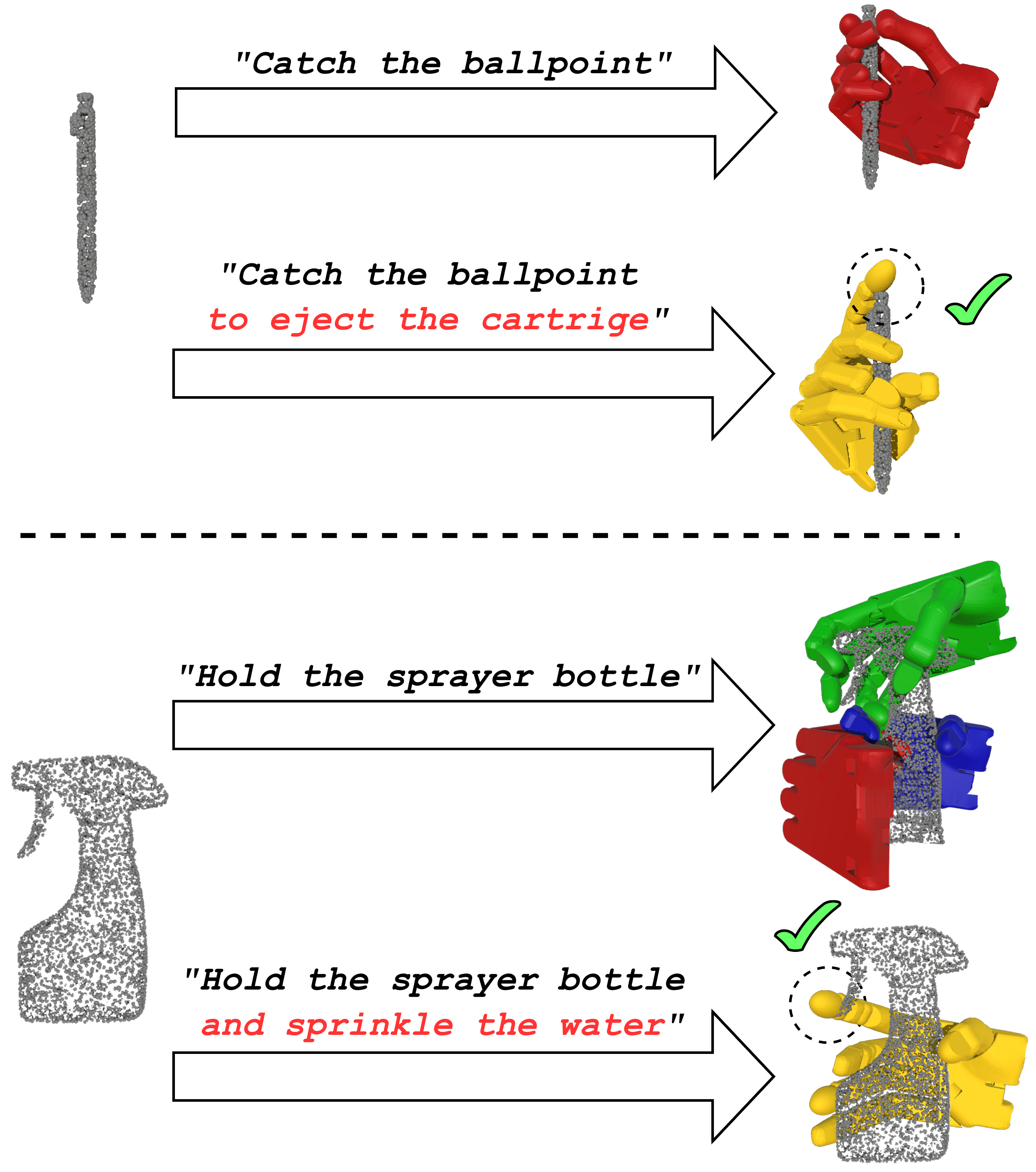}
    \caption{\textbf{Task-oriented grasp.} The task-agnostic grasp only ensures the grasp is stable, while the task-oriented grasp needs to contact the affordance part for the downstream tasks.}
    \label{fig:tog}
\end{figure}

Grasping is the first step to accomplishing generic prehensile manipulation tasks. In common manipulation scenarios, humans execute grasping with a specific task intention, facilitating the grasp selection and minimizing the need for repeated re-grasping \cite{regrasp}. Grasping concerning downstream tasks is termed ``task-oriented grasping'' (TOG). As shown in Fig. \ref{fig:tog}, unlike conventional grasping tasks \cite{grasp_stable}, which solely aim to achieve stable object picking without considering the purpose of the grasp,
TOG tries to find the optimal grasp pose to execute manipulation tasks directly.
Previous works on TOG \cite{2ftog_data,graspgpt} have predominantly focused on 2-finger parallel grippers, which offer limited dexterity and constrain the complexity of achievable tasks. In contrast, TOG with a dexterous hand, a more generic manipulation setting, is seldom explored \cite{dextog_language}.

The challenges to design a dexterous task-oriented grasping prediction learning framework have three folds: (1) \textbf{Task constraint.} The system should understand the task and make it a constraint for grasp planning; (2) \textbf{Multiple valid grasps.} Given the task constraint, the potential optimal grasps on the target objects are non-unique. The system should support such a multi-modal distribution of grasp poses. (3) \textbf{High degrees of freedom (DoF).} Unlike grasping with a 2-finger parallel jaw gripper, dexterous grasping should search for a valid grasp pose in high-dof configuration space and consider force stability. To address these challenges, we propose a novel language-guided diffusion-based learning framework, \textbf{DiffuTOG}. Given a prehensile manipulation task, DiffuTOG directly predicts the grasp pose by iteratively adding noises and denoising in hand configuration space, with natural language task description, 3D object observation, and hand model as the conditions. These conditions are embedded with different encoders.

To train DiffuTOG, we need a dataset concerning multiple object categories and diverse task settings. However, we find no existing dexterous manipulation datasets are suitable. 
Therefore, we build a data engine for dexterous TOG task, named \textbf{DexTOG}. It works in a coarse-to-fine, sparse-to-dense manner. First, we generate task-agnostic grasp poses around the objects with a given hand and object models. For each task, we apply heuristic rules to coarsely filter out the task-relevant grasp poses. Then, we utilize the filtered grasp poses to train DiffuTOG. Due to the multi-modal nature of the diffusion model, DiffuTOG can amplify the relevant grasp poses near the object. However,
the task-relevant grasps that are filtered by heuristic rules and amplified by DiffuTOG may still not align perfectly with the subsequent task. Therefore, we adopt goal-conditioned reinforcement learning to validate the task-relevant grasps. The successfully executed grasps are the final task-oriented grasps. During dataset construction, the DiffuTOG and reinforcement learning policy are pre-trained, and the task-oriented grasp poses are automatically labeled.

With DexTOG, we efficiently build a dataset, DexTOG-80K. It consists of 80K grasps on 80 objects from 5 categories with respect to 5 tasks performed by a Shadow robot hand: \textbf{Stapler clicking}, \textbf{Spray bottle pressing}, \textbf{Spray bottle triggering}, \textbf{Bottle cap twisting} and \textbf{Ballpoint pen pressing} on these objects. These tasks are designed to utilize the dexterity of the Shadow robot hand. The task is described in natural language and contains information about ``action'', ``target'' and ``task''. 

To evaluate the model, since many dexterous task-oriented grasp methods are not open-source yet, we adapt two task-agnostic dexterous grasp methods, GraspTTA \cite{grasptta}, Unidexgrasp \cite{unidexgrasp} to TOG setting, denoted as GraspTTA-TOG and Unidexgrasp-TOG. We conduct both task-agnostic and task-oriented grasp planning. The extensive experiments show that our method outperforms the baseline methods.

We conclude our contributions as follows:
\begin{itemize}
    \item DiffuTOG. A diffusion-based dexterous grasp generation method for both task-agnostic and task-oriented tasks, based on textual task description.
    \item DexTOG. A data engine to generate large dexterous dataset with heuristic rules and RL-based policy. DiffuTOG is the core component for pose augmentation.
    \item DexTOG-80K. A dataset generated by DexTOG. It consists of text labeled task-oriented and task-agnostic dexterous grasp poses. There are 80K shadow grasp poses on 80 articulated objects.
\end{itemize}

\section{Related Works}
\ralhigh{
Our work is most related to those methods focusing on task-oriented grasp pose planning and dataset generation.}
\subsection{Task-Oriented Grasp Prediction}
\ralhigh{
Task-oriented grasping is a special grasp pose planning task, which not only consider the stability of an object's grasp but also the constraints of specific tasks.
}

\ralhigh{
\textbf{Parallel Grasping}
Prior studies have primarily focused on 2-finger robot grippers, where the grasp pose is typically characterized by a 6-D pose.
Detry et al. \cite{2ftog_afford} pioneered the use of affordance areas to associate stable grasps with downstream tasks, followed by numerous studies \cite{2ftog_crf, 2ftog_activity, 2ftog_clip, phygrasp, 2ftog_lerf}.
These methods filter the desired, task-oriented grasps from the base grasp detector results by judging whether the contact points are located within the affordance area. However, it's not enough to consider the contact point to fit the subsequent tasks. A more detailed attribution of the subsequent task should be taken into account. 
However, to adequately prepare for subsequent tasks, it is insufficient to consider only the contact points.
Pantankar et al. \cite{2ftog_skew} introduced the concept of task skew, which only applies to objects with regular shapes, such as boxes and cylinders.
Recent advancement of large language models makes it possible to encode more complex task constraints into TOG pipelines\cite{graspgpt, 2ftog_clip, phygrasp, 2ftog_lerf}. Tang et al. \cite{2ftog_clip} pioneered this approach by  GraspCLIP, which depends on a 2D vision-language model, making it only work under 2D-like, top-down grasping scenarios. Later, they extend this method into GraspGPT~\cite{graspgpt}, which can generate 6D grasp poses powered by 3D vision-language encoders.}

\ralhigh{
\textbf{Dexterous Grasping}
The limitations of parallel grippers restrict tasks to simple manipulation operations with limited coverage of everyday activities. On the other hand, dexterous hands feature a high-dimensional configuration space for the grasp pose, leading to more complex requirements for subsequent tasks.
Previous studies on dexterous task-oriented grasping have primarily focused on mimicking trajectories of objects \cite{dextog_human}, contact points \cite{dex_pregrasp}, or key points of the hand \cite{dextog_language} derived from human demonstrations.
However, these approaches generally treat the object as a rigid body, which often fails to fully leverage the capabilities of dexterous hands in performing in-hand manipulations. In comparison, the proposed DexTOG framework tries to focus on the articulated objects. 
}

\subsection{Dataset of Dexterous Manipulation}
\ralhigh{
Data-driven grasp methods heavily rely on large-scale datasets.
Most existing datasets for manipulation focus on 2-finger parallel grasping \cite{2ftog_data,2ftog_clip,acronym} and human grasping \cite{hoi1,dexycb,oakink,hoi2,hoi3}.
Building dexterous grasp datasets usually involves much more time and human labor.
Parallel computing techniques are widely used to accelerate data collection with differentiable optimization frameworks \cite{ddg,gendexgrasp,unidexgrasp,dexgraspnet}.
However, the differentiable optimization objective functions are usually adapted from some grasp metrics \cite{grasp_stable}, which can facilitate the generation of task-agnostic grasp poses given object models but can not generate task-oriented grasp poses.
To solve this problem, our work introduces a closed-loop data engine designed to generate and verify task-oriented grasps autonomously. This innovative approach not only enhances the efficiency of data generation but also improves the accuracy and reliability of the grasps for complex manipulative tasks.
}

\section{DiffuTOG}
\begin{figure}
    \centering
    \includegraphics[width=\linewidth]{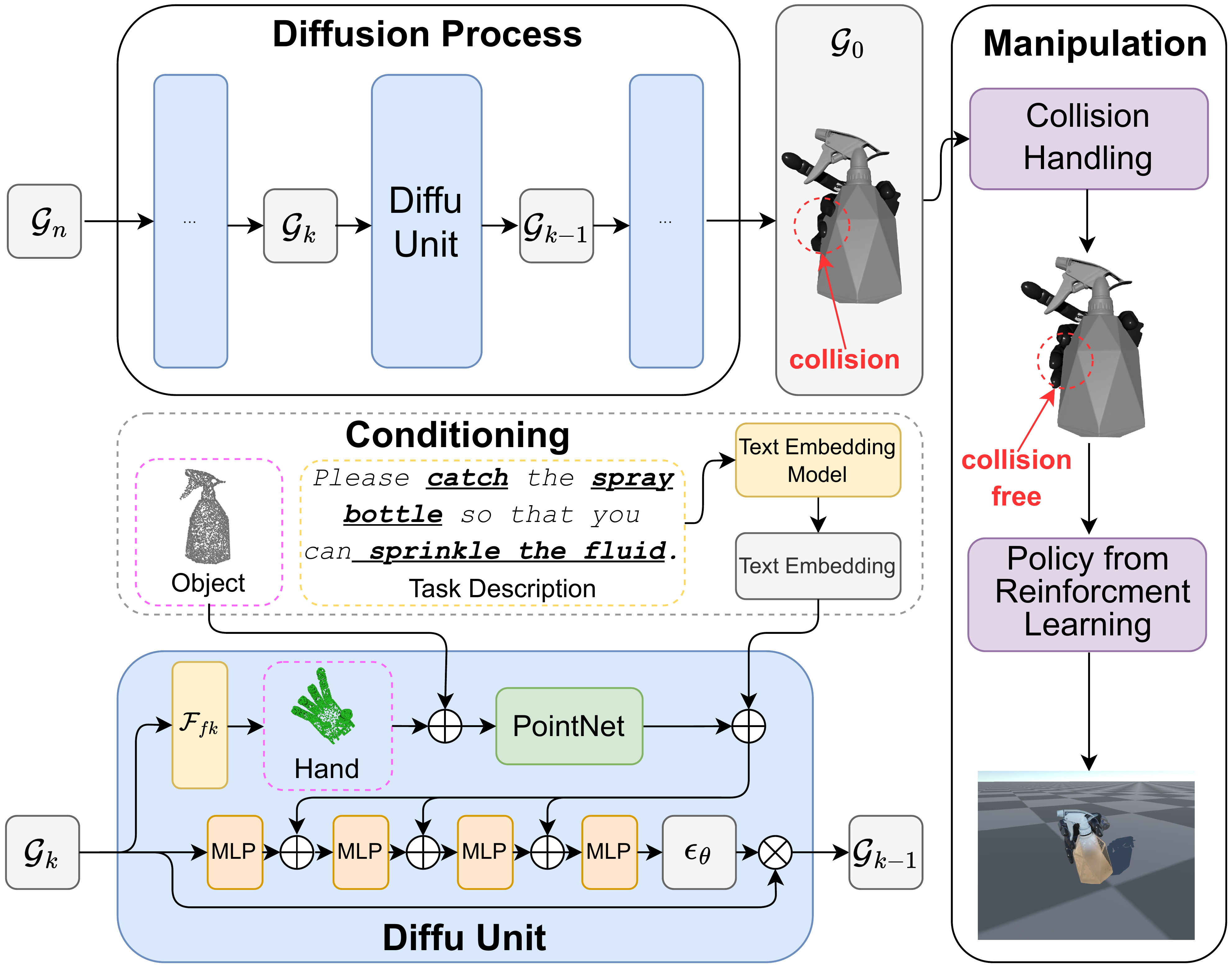}
    \caption{\textbf{Pipeline.} Our method contains two stages: grasp generation and grasp execution. In the generation stage, DiffuTOG generates grasp proposals, and then a test-time optimizer is used to refine the proposals. In the execution stage, we use the refined grasp pose as the initial pose and train the state-based RL to complete the task.
    The execution stage here is only for verification purposes.
    }
    \label{fig:pipeline}
\end{figure}

In this section, we describe the design and training of DiffuTOG. 
Given a 3D observation of an object, $\mathcal{O}\in \mathbb{R}^{N_1\times 3}$, a task description $\mathcal{T}$ and the robot hand model $\mathcal{M}$ with a configuration space of $\mathcal{G}$, DiffuTOG tries to predict a grasp pose $\mathcal{G}_k=(R_k,t_k,q_k) \in \mathcal{G}$ in a denoising diffusion process. $N_1$ is the point number of the observed point cloud, $k$ is the iteration index, $R\in SO(3)$ represents the wrist rotation, $t\in \mathbb{R}^3$ means the translation, $q\in \mathbb{R}^{J}$ is the joint pose for a $J$-DoF dexterous hand. 
For each robot hand pose $\mathcal{G}_k$, we can obtain the robot hand 3D point cloud $\mathcal{H}\in \mathbb{R}^{N_2\times 3}$ with a forward kinematics function $\mathcal{F}_{fk}$, $\mathcal{H}_k = \mathcal{F}_{fk}(\mathcal{G}_k)$.
\ralhigh{The overall framework is illustrated in Fig. \ref{fig:pipeline}.}


\subsection{Grasping Generation Through Denoising Diffusion Probabilistic Models (DDPMs)}
We formulate the task-agnostic grasping generation as an unconditioned diffusion process \cite{ddpm}, and thus, the task-oriented grasp can be regarded as a conditional diffusion process.

In a typical diffusion model setup, the denoising process from $\mathcal{G}_T$ predicts the desired dexterous grasp $\mathcal{G}_0$. The iterative equation is as follows:

\begin{equation}\label{eq:ddpm}
    \mathcal{G}_{k-1} = \frac{1}{\sqrt{\alpha_t}}(\mathcal{G}_k - \frac{1 - \alpha_t}{\sqrt{1 - \bar{\alpha}_t}} \epsilon_\theta (\mathcal{G}_k, k) ) +  \mathcal{N}(0, \sigma^2I),
\end{equation}
where $\epsilon_\theta$ is the noise prediction network with parameters $\theta$ that will be optimized through learning and $\mathcal{N}(0, \sigma^2I)$ is Gaussian noise added at each iteration. 

\paragraph{Grasp Pose for Diffusion Model} To represent the wrist rotation, the quaternion vector is widely adopted \cite{dex_pregrasp,dextog_human,dextog_language}. However, rotation in quaternions is applied by multiplication, and the noise in the diffusion process is set by addition. That is, the increment of the quaternion does not correspond to the increment of rotation. Thus, a slight noise in the final output could lead to a meaningless pose, which makes the learning unstable.

To address this issue, we employ and adapt the 6D rotation representation as suggested in \cite{rotation_representation}. This allows us to construct a unique rotation matrix $R$ for two arbitrary 3D-vector $\mathbf{p_1} = [x_1, x_2, x_3]$, $\mathbf{p_2} = [x_4, x_5, x_6]$. If $\mathbf{p_1}$ and $\mathbf{p_2}$ can be orthogonalized as follow:
\begin{align}
    \mathbf{r_1} &= \frac{\mathbf{p_1}}{\Vert \mathbf{p_1} \Vert}, \mathbf{r_2} = \frac{\mathbf{p_2} - \mathbf{p_1} \cdot \mathbf{p_2}}{\Vert \mathbf{p_2} - \mathbf{p_1} \cdot \mathbf{p_2} \Vert}, \\
    R &= [\mathbf{r_1^T}, \mathbf{r_2^T}, \mathbf{(r_1 \times r_2)^T}],
\end{align}
we can represent the hand-wrist coordinate with $R$, which can be alternatively represented by $X = [\mathbf{p_1}, \mathbf{p_2}]$. In this representation, a noise in rotation can be denoted as $X_{\epsilon} \in \mathbb{R}^6$ and is assumed to follow a standard normal distribution $\mathcal{N}(0,I)$. It is noteworthy that the addition of this noise corresponds to an increment in rotations.

\subsection{Conditional Embedding}
We add three conditional embeddings during the diffusion process to regularize the grasp and adapt it to specific objects and task descriptions.

\paragraph{Hand Encoder} To encode the hand geometry, we first augment the reconstructed hand point cloud $\mathcal{H}_k \in \mathbb{R}^{N_2\times 3}$ with full-1 vector, and have $\mathcal{H}^\prime_k \in \mathbb{R}^{N_2\times 4}$. Then, we put it through a PointNet \cite{pointnet} and result in a 128-d vector, $Emb_{H} \in \mathbb{R}^{128}$.

\paragraph{Object Encoder} Similarly, we augment the object point cloud $\mathcal{O}\in \mathbb{R}^{N_1\times 3}$ with a full-0 vector, and produce $\mathcal{O}^\prime\in \mathbb{R}^{N_1 \times 4}$. $\mathcal{O}^\prime$ is also encoded by the PointNet and resulted in a 128-d vector $Emb_{O} \in \mathbb{R}^{128}$. Unlike the hand encoder, since the object observation is unchanged during the denoise iteration, the object encoder will be called only once.

\paragraph{Task Description Encoder}
To make the output grasp pose task-aware, we use task description embedding as the guidance. Given a task description, $\mathcal{T}$, we first use the OpenAI embedding model, \textit{text-embedding-ada-002} \cite{ada} to obtain a text embedding $Emb^\prime_{T} \in \mathbb{R}^{1536}$. Then, we use 3-layer MLP to compress the embedding into $Emb_{T} \in \mathbb{R}^{256}$.

\subsection{Conditional DDPM Training}
After getting conditional embeddings, we directly concatenate them with the grasp pose feature and pass through each MLP in each Diffu Unit as illustrated in Fig. \ref{fig:pipeline}. A ``Diffu Unit'' is a conditional decoder that consists of multiple MLPs to predict $\epsilon_\theta$. 

Finally, the integration of conditions namely object observation $\mathcal{O}$ and task description $\mathcal{T}$, and the hand model $\mathcal{M}_k$ makes the original DDPM in Eq. \ref{eq:ddpm} to a conditional DDPM:

\begin{equation}\label{eq:cddpm}
\mathcal{G}_{k-1} = \frac{1}{\sqrt{\alpha_t}}(\mathcal{G}_k - \frac{1 - \alpha_t}{\sqrt{1 - \bar{\alpha}_t}} \epsilon_\theta (\mathcal{G}_k, \mathcal{O}, \mathcal{T}, \mathcal{M}_k, k) ) +  \mathcal{N}(0, \sigma^2I).
\end{equation}


The learning process can be conducted by the diffusion loss term:
\begin{equation}
    L_{D} = \mathbb{E}_{\mathcal{G},\mathcal{O}, \mathcal{T}, k, \epsilon \sim \mathcal{N}(0,I)}[\Vert \epsilon - \epsilon_\theta(\mathcal{G},\mathcal{O}, \mathcal{T}, \mathcal{M}_k, k) \Vert^2],
\end{equation}
where $\epsilon$ is the added Gaussian noise, and $\epsilon_\theta$ is the estimated noise.

In addition, we use a reconstruction loss on the robot hand model $\mathcal{H}_k$. The dense supervision can make the training stable.
\begin{align}
    \mathcal{G}^\prime &= \sqrt{1 - \beta_k} \mathcal{G} + \sqrt{ \beta_k}\epsilon, \\
    \mathcal{G}^\prime_\theta &= \sqrt{1 - \beta_k} \mathcal{G} + \sqrt{\beta_k}\epsilon_\theta, \\
    L_{R} &= \mathbb{E}[\Vert \mathcal{F}_{fk}(\mathcal{G}^\prime) - \mathcal{F}_{fk}(\mathcal{G}^\prime_\theta) \Vert].
\end{align}
    In summary, our overall loss function is:
    \begin{equation}
        L = L_D + \lambda_R L_R,
    \end{equation}
where $\lambda_R$ is the weighting coefficient, which will be determined by cross-validation.

\subsection{Test-Time Collision Handling}
At the test time, the grasp pose is generated by DiffuTOG from random noise. However, since the denoising process does not guarantee collision-free grasps, the generated grasp pose might be in a collision. To mitigate this issue, we adjust the imperfect grasp pose by minimizing the penetration energy $E_{\text{pene}}$ using gradient descent:
\begin{align}
    E_{\text{pene}} &= \max \{ \max_{x \in \mathcal{H}} \sigma(x, \mathcal{O}),  \max_{x \in \mathcal{O}} \sigma(x, \mathcal{H}) \}, \\
    \sigma(u, \mathcal{M}) &= \left \{ 
    \begin{aligned}
        d,  & &\text{if $u$ inside $\mathcal{M}$}; \\
        0, & & \text{otherwise}.
    \end{aligned} \right.
\end{align}
where $\mathcal{O}$ is the mesh of object, $\mathcal{H}$ is the mesh of hand, $d$ is the distance from $u$ to the surface of $\mathcal{M}$. By decreasing the energy, the process tries to pull the hand's deepest point inside the object out of it.

%
\section{DexTOG, Data Engine}
\begin{figure}
    \centering
    \includegraphics[width=0.85\linewidth]{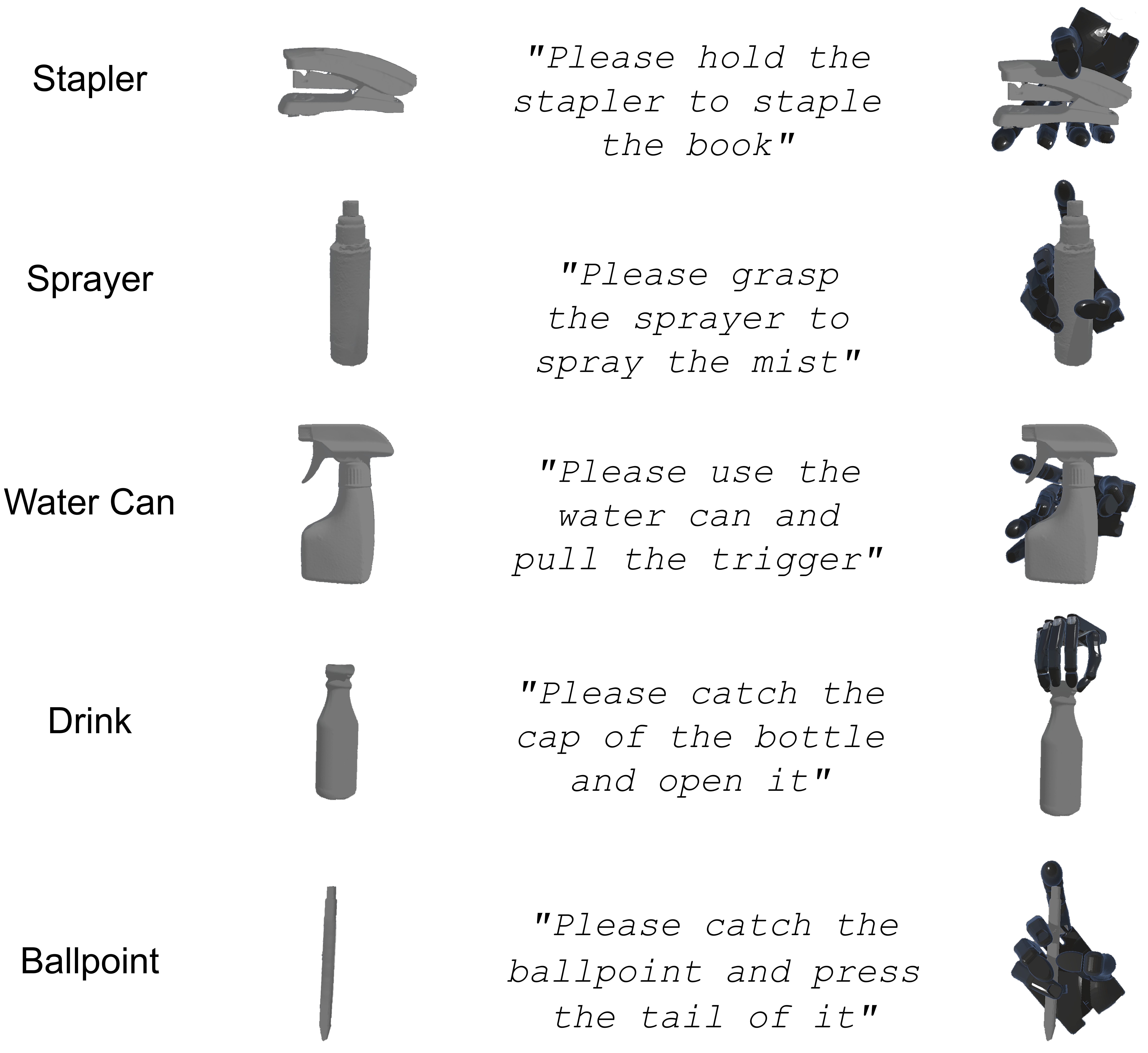}
    \caption{\textbf{Samples in DexTOG-80K.} The object and the corresponding task-oriented grasp.}
    \label{fig:dataset_samples}
\end{figure}

In this section, we first define a few exemplar tasks and then describe how to generate task descriptions with natural language for these tasks. Next, we introduce the data engine DexTOG to produce training data for learning task-oriented grasping. Finally, we report statistics of the generated TOG dataset, DexTOG-80K. \ralhigh{Fig. \ref{fig:dataset_samples} shows some samples in the dataset.}

\subsection{Task Definition} 
\label{sec:task-def}
We define five tasks that involve interaction with articulated objects: stapler clicking, spray bottle pressing, spray bottle triggering, bottle cap twisting, and ballpoint pen pressing. The reason why we choose articulated objects is that articulation provides more DoFs, and thus they can benchmark grasps for many meaningful in-hand manipulation tasks.

\begin{itemize} 
\item \textbf{Stapler Clicking}: A stapler is grabbed so that the stapling can be done by pushing between the thumb and other fingers. 
\item \textbf{Spray Bottle Pressing}: A spray bottle is grabbed, and one of the fingers (ideally the index or middle finger) is ready to press the button while the remaining four fingers grasp the sprayer bottle stably.
\item \textbf{Spray Bottle Triggering}: A spray bottle is grabbed, and one of the fingers (ideally the index or middle finger) is ready to pull the trigger while the remaining four fingers grasp the sprayer bottle stably.
\item \textbf{Bottle Cap Twisting}: The bottle cap is in contact and is about to be opened. The bottle is assumed to be fixed.
\item \textbf{Ballpoint Pen Pressing}: A ballpoint pen is held, and one of the fingers (ideally the thumb or index finger) is ready to press the button.
\end{itemize}

These tasks are notably challenging for a dexterous robotic hand, as they require not just a stable grip but also precise finger placement near specific functional components, often referred to as affordance parts.


\subsection{Task Description Generation}
Building upon the framework proposed by \cite{langshape}, we design templates and attributes to generate textual task conditions systematically. 
A typical template is built upon a triplet (\textit{action}, \textit{part}, \textit{affordance}) by filling several conjunction words between the elements. For example, \textit{please $\langle$action$\rangle$ the $\langle$part$\rangle$ so that you can $\langle$affordance$\rangle$}.
The choices for \textit{$\langle$action$\rangle$} could be: \textit{grasp, catch}. The choices for \textit{$\langle$part$\rangle$} could be: \textit{cap, top, cap of the bottle}. And the choices for \textit{$\langle$affordance$\rangle$} could be: \textit{open the bottle, drink the water, twist it}.
ChatGPT generates the conjunction words by iteratively asking ``Please compose the words \textit{$\langle$action$\rangle$}, \textit{$\langle$part$\rangle$}, \textit{$\langle$affordance$\rangle$} to generate a sentence for task description''. In this way, the template can be more natural and diverse.

To note, the same workflow can be applied to generate text conditions for task-agnostic grasping by simply removing the \textit{affordance} element in the triplet.

In total, we have $22$ templates, and $5$ action attributes, $16$ part attributes, $18$ affordance attributes.
A full list of templates and attributes can be referred to in the supplementary materials.

\subsection{Task-oriented Grasp Generation}
To pair the task description with a grasp pose, we randomly select the object instance from the AKB-48 dataset \cite{akb48}.
For task-agnostic grasping, we employ an analytical grasp planning algorithm, ISF \cite{isf}, to generate the grasp pose. The planned grasp pose is validated in a physics-based simulator \cite{rfu}.
For task-oriented grasping, we start with the task-agnostic grasp poses and perform a coarse-to-fine pipeline to filter out the task-oriented grasp poses from the task-agnostic ones.

The task-agnostic grasp poses produced by ISF have a good coverage rate over the object surface. However, since they do not consider task constraints, only nearly 0.1\% of them are valid for the downstream tasks. Thus, finding the 0.1\% part and amplifying the quantity of valid TOG are two critical challenges in building the dataset.

Previous works usually adopt an intermediate representation on objects, affordance map \cite{langshape, phygrasp}, to help filter the task-oriented grasps. However, obtaining an accurate affordance map for different tasks requires training affordance prediction networks, which also need training data. Besides, the affordance maps on object surfaces do not guarantee a valid grasp. 

\begin{figure}
    \centering
    \includegraphics[width=0.9\linewidth]{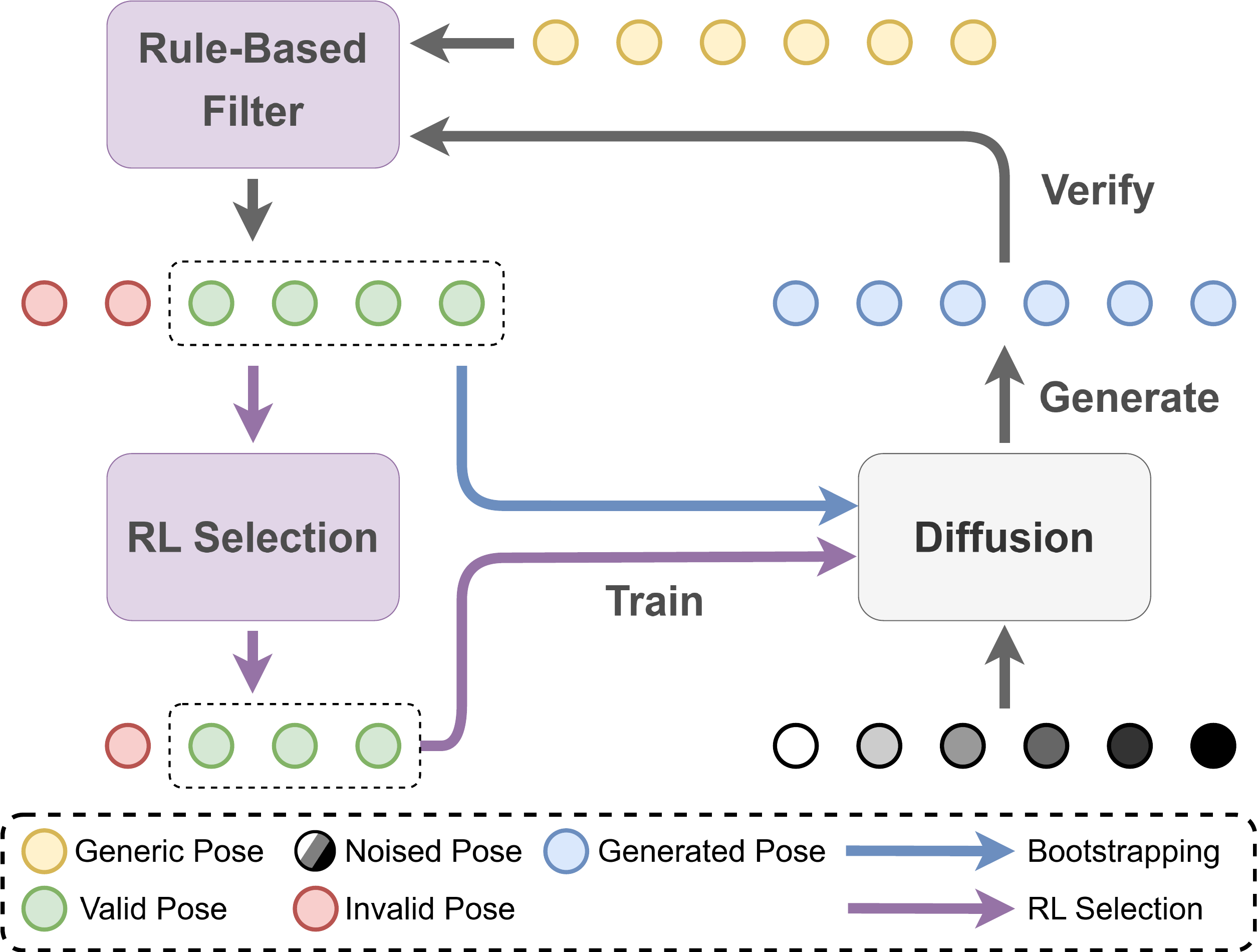}
    \caption{\textbf{Data generation process.} The generic poses, which are the task-agnostic grasp poses, are first filtered by some rules. Then, the rule-filtered grasps are sent to DiffuTOG to amplify the grasp quantity. The RL policy finally verifies the amplified grasp poses.}
    \label{fig:datageneration}
\end{figure}

Instead, we follow a generic-to-specific, coarse-to-fine path. The whole data generation process is shown in Figure.~\ref{fig:datageneration}

\subsubsection{Rule-based Filtering}
Initially, we apply heuristic criteria to discern the subset of valid grasps as follows: 

Mark the fingertip point $i$ as $p^{i}$, the normal of fingertip pad as $n^{i}$, such as $p^{index}$ and $n^{index}$. And we use $d(x, y)$ to mark the minimum Euclidean distance between two surfaces $x$ and $y$.
\begin{itemize}
    \item \textbf{Stapler Clicking.} Mark the top surface of the stapler as $S^{top}$, the bottom surface of the stapler as $S^{bottom}$, the corresponding average normal as $n^{top}$ and $n^{bottom}$, we only keep the grasp pose satisfying:
    \begin{equation}
    \left \{
    \begin{aligned}
        &\max\{d(p^{thumb}, S^{top}), d(p^{i}, S^{bottom}) \} \leq 5mm, \\
        & n^{thumb} \cdot n^{top} < 0, \\
        & n^{i} \cdot n^{bottom} < 0,
    \end{aligned}
    \right.
    \end{equation}
    where $i \in \{ index, middle\}$, or reverse the position of thumb and index/middle.
    \item \textbf{Spray Bottle Pressing.} Mark the top surface of button as $S^{button}$, we only keep the grasp pose satisfying:
    \begin{equation}
    \min\{d(p^{thumb}, S^{button}), d(p^{index}, S^{button}) \} \leq 5mm.
    \end{equation}
    \item \textbf{Spray bottle Trigger.} Mark the outer surface of the trigger as $S^{trigger}$, the average normal of the surface as $n^{trigger}$, we only keep the grasp pose satisfying:
    \begin{equation}
        d(p^{index}, S^{trigger}) \leq 5mm, \ n^{index} \cdot n^{trigger} < 0.
    \end{equation}
    \item \textbf{Bottle Cap Twisting.} Mark the center of the bottle cap as $c_{cap}$, we only keep the grasp pose satisfying:
    \begin{equation}
    \forall i, \  d(p_i, c_{cap}) \leq 2.5cm.
    \end{equation}
    \item \textbf{Ballpoint Pen Pressing.} Mark the top surface of button as $S^{button}$, we only keep the grasp pose satisfying:
    \begin{equation}
    \min\{d(p^{thumb}, S^{button}), d(p^{index}, S^{button}) \} \leq 2.5mm.
    \end{equation}
\end{itemize}

These rules do not accurately correspond to task-oriented affordance, but they require no training and cover the affordance part coarsely. After this step, around 0.1\% of grasps can be retained, which is few for a dataset. Thus, we need to amplify the quantities. To achieve this, we propose to bootstrap with DiffuTOG.

\subsubsection{Bootstrapping with DiffuTOG}
We use these filtered grasp with the corresponding task description to train the DiffuTOG model. In the training phase, we randomly sample $10240$ random grasp poses, input to the denoising process. The denoised poses are fed back into the system as input for the next iteration. Through successive iterations, this process effectively amplifies the limited instances of valid grasp poses to a sufficient quantity.

In this stage, the DiffuTOG is trained with rule-selected data, which are coarsely aligned with the task affordance area. Thus, the amplified data cannot be guaranteed to match the task description. However, the density of grasps near the rule-selected areas is largely amplified.
Finally, we use the amplified data to train a reinforcement learning policy. All the grasp poses that can support accomplishing the tasks are the final task-oriented grasp poses.

\subsubsection{TOG Selection with Reinforcement Learning}
We validate the amplified grasp poses in a simulator \cite{rfu}. We adopt Proximal Policy Optimization (PPO) to train a policy for each task.
The design of the reward function encourages the use of task-oriented grasps generated by DiffuTOG:

\begin{itemize}
\item \textbf{Task Reward} ($r_t$): A positive reward is granted upon the process of a specified task. Examples of such tasks include expelling a staple, pressing a sprayer, or similar actions. This reward motivates the learning agent to achieve the ultimate goal of the task at hand, reinforcing behaviors that lead directly to task accomplishment.
\begin{equation}
    r_t = \alpha_1 (\theta - \theta_0),
\end{equation}
where the $\theta$ denotes the normalized angle of the articulated object joint.

\item \textbf{Lift Reward} ($r_l$): A positive reward motivates the policy to lift the object. Such reward is designed to check if the grasp pose is stable enough to hold the object:
\begin{equation}
    r_l = \alpha_2 \min \{(h - h_0), h_{\text{max}}\},
\end{equation}
where $h$ is the height of the object center, the $h_0$ is the initial height of the object center, and the $h_{\text{max}}$ is a threshold to avoid the policy only learning to lift the object higher.

\item \textbf{Task Completion} ($r_c$): A large positive constant to reward if the task is completed:
\begin{equation}
    r_c = \alpha_3 [h > \hat{h}, \theta > \hat{\theta}],
\end{equation}
where $\hat{h}$ and $\hat{\theta}$ is the pass line for object height and joint angle, $[\cdot]$ is the condition function that equals to $1$ if the condition in bracket is satisfied else $0$.

\item \textbf{Drop Penalty} ($p_d$): To avoid squeezing the object away, we add a penalty to punish the action that pushes the object away from the hand:
\begin{equation}
p_d = -\alpha_4 \text{ dist}(t_{\text{hand}}, t_{\text{object}}),
\end{equation}
where $t_{\text{hand}}$ denotes the position of hand palm center and $t_{\text{object}}$ denotes the position of object center.
\end{itemize}
The total reward is:
\begin{equation}
r = r_t + r_l + r_c + p_d.
\end{equation}
By actually executing the generated grasps, the RL policy network filtered the grasps in the possible affordance areas. After the RL policy network's judgment, we aggregate the successful task-oriented grasp to fine-tune the DiffuTOG further.

\subsection{Data Generation and Statistics}
With the generic-to-specific, coarse-to-fine TOG generation loop, we can obtain theoretically endless valid grasp poses and a well-trained DiffuTOG and RL policy network.

In this work, we choose 80 objects from the AKB-48 dataset \cite{akb48} and generate over 400K valid grasp in total. We finally sampled 1k valid grasp per object, 500 task-oriented grasp, and 500 task-agnostic grasp, which got 80K grasp poses in total. The reason why we keep a task-agnostic grasp is that they are more diverse in both wrist poses and joint states, which can also help the training of neural networks. Besides, in this way, the dataset can support both the TOG Task and the task-agnostic grasp task. We randomly split the dataset into seen and unseen objects with a ratio of 9:1 and trained our data on the seen objects.


\section{Experiment}
\subsection{Experimental Setup}
\subsubsection{Simulation} We utilized RFUniverse~\cite{rfu} as our simulation environment. In our experimental setup, the mass of each object was set to 300g, except for the ballpoint pen, which was set to 100g. 

In the rule-based filtering process, we check the task-agnostic grasp poses by lifting the object 5cm and applying gravity to the object to see if it will fall.
In the RL training, we placed objects such as the sprayer, bottle, and water can upright on the ground. Conversely, the stapler and ballpoint were positioned randomly on the ground. We want the objects to be placed on the table in the most common way possible.

\subsubsection{Implementation Details}
 We train our model on one NVIDIA A40 with 300 epochs, using $\lambda_R = 1$. In the inference stage, we optimize the generated data with 200 steps to handle collision. Then, we use the PPO \cite{ppo} as our policy algorithm to accomplish the tasks we set out. For the reward function, we set $\alpha_1 = 80, \alpha_2 = 10, \alpha_3 = 50, \alpha_4 = 10, h_{\text{max}} = 15cm, \hat{h} = 10cm, \hat{\theta} = 0.6$. And we train ppo with $2,000,000$ iterations, learning rate $10^{-4}$, horizon $100$.

\subsubsection{Metrics}
We use three metrics to measure our approach:
\begin{itemize}
    \item $Q_1$~\cite{grasp_stable}: The smallest wrench needs to make a grasp unstable. This indicates how stable the grasp is in the aspect of force closure. To avoid grasp with large penetration interfering with the quality, we record the grasp with penetration bigger than 0.5cm as zero.
    \item Object penetration depth (cm): The maximal penetration from the object point cloud to hand mesh.
    \item Success rate: The success cases among all generated grasp posed via the RL policy trained on the dataset. 
\end{itemize}

\subsection{Result of Task-Agnostic Grasp}
As a specialized form of task-oriented grasp, task-agnostic grasp (the task can be regarded as ``lift and hold'') is more appealing in the community. We also benchmark our method on the task-agnostic grasp task for a broader audience.

We compare our method with two baselines, GraspTTA~\cite{grasptta} and UniDexGrasp~\cite{unidexgrasp} on DexTOG-80K. The training of GraspTTA and UniDexGrasp maintains the same setting as the original works. The result is shown in Table. \ref{tlb:generalgrasp}.


\begin{table}
\centering
\caption{\textbf{Quantitative result of task-agnostic grasp}. pen: object penetration (cm) }
\label{tlb:generalgrasp}
\setlength{\tabcolsep}{6pt}
\begin{tabular}{ccccc}
\toprule
\multirow{2}{*}{Method} & \multicolumn{2}{c}{seen obj}         & \multicolumn{2}{c}{unseen obj} \\ \cline{2-5} 
                        & $Q_1 \uparrow$ & pen$\downarrow$  & $Q_1 \uparrow$        & pen $\downarrow$        \\ \hline
GraspTTA                &  0.0391     &   0.832  &   0.0178     &   0.923    \\
UniDexGrasp             &  0.0734   &  \textbf{0.201}   &  0.0698    &  \textbf{0.255}     \\
Ours                    &   \textbf{0.1067}    &     0.410       &      \textbf{0.0930}        &  0.385        \\ \bottomrule 
\end{tabular}
\end{table}

\subsection{Results of Task-Oriented Grasp}
\begin{figure}
    \centering
    \includegraphics[width=0.8\linewidth]{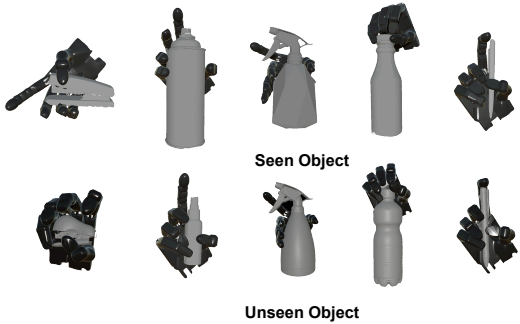}
    \caption{\textbf{Qualitative Results} of the generated task-oriented grasp on both seen and unseen objects.}
    \label{fig:qualityresult}
\end{figure}

\ralhigh{
The RL policy is trained on the proposed DexTOG-80K dataset.
Then, we filter the collision-free grasp poses generated from the DiffuTOG model and use them as the initial poses.
The GraspTTA is a popular baseline in this track. To note, the original GraspTTA adopts a C-VAE to control the grasp type. Here, for a fair comparison, we modify it to use the text embeddings instead.
The qualitative result of the task-oriented grasp generated by our method is shown in Figure \ref{fig:qualityresult}. The quantitative results are shown in Table \ref{tlb:taskorientedgrasp}.
Results show that the proposed method outperforms the baseline. The reason why GraspTTA does not perform well is that it tends to generate grasp poses that may collide with and even penetrate into objects. Also, it may ignore the task condition, producing some grasp poses that are not good for downstream manipulation tasks.}

\ralhigh{It's also worth noticing that the success rates differ across various tasks.
For example, the success rate for a sprayer trigger is much higher than the one for a ballpoint pen.
The reason is that the current methods assume the objects stay static during the grasp process. This is true for the sprayer trigger since pressing a sprayer trigger tends to hold the sprayer stably.
However, due to its light weight, the action of the fingers can dramatically shift the position of the ballpoint pen. To be specific, there is about a 57\% chance of the ballpoint pen shifting more than 1 cm after being picked up from the ground. Considering the pen's button is typically smaller than 1 cm, the shift can significantly impact the success rate of the manipulation task.
These differences in success rates highlight the importance of considering the relationship between the grasp and the task action to achieve a good task-oriented grasp.
}

\begin{table}
\centering
\caption{\textbf{Success Rate of task-oriented grasp using the RL policy.} DexTOG-80k: The success rate of the dataset grasp. GraspTTA$^*$: The modified GraspTTA.}
\label{tlb:taskorientedgrasp}
\setlength{\tabcolsep}{2pt}
\begin{tabular}{cccccc}
\toprule
Task                & stapler & sprayer press & sprayer trigger & bottle & ballpoint \\ \midrule
DexTOG-80k &    $85\%$  & $57\%$ & $70\%$ &  $81\%$& $26\%$ \\ \midrule
GraspTTA$^*$(seen) &    $10\%$   &  $12\%$ & $0\%$  & $0\% $ & $12\%$ \\
Ours (seen) &   $67\%$  & $28\%$ & $54\%$  &  $73\%$& $19\%$ \\ \midrule
GraspTTA$^*$(unseen) &    $4\%$   &  $9\%$ & $0\%$  & $0\% $ & $3\%$ \\
Ours (unseen) &   $46\%$  & $14\%$ & $27\%$ & $41\%$ & $15\%$ \\ 
\bottomrule
\end{tabular}
\end{table}

\subsection{Ablation Study}
\subsubsection{Hand Point Cloud Encoder}
Unlike other diffusion-based dexterous grasp generation methods \cite{dexdiffuser, ugg}, which generate task-agnostic grasps without hand geometry, we include a hand geometry encoder in our method. To demonstrate its importance, we compare the quality of grasps between the model with \( \text{Emb}_{H} \) and the one without \( \text{Emb}_{H} \). The results, shown in Table \ref{tlb:tta}, indicate that integrating the hand geometry encoder improves grasp performance.

\subsubsection {Test-time Collision Handling}
We compared the quality of grasp output directly from DiffuTOG and after Test-time collision handling (``+ T''); the result is shown in Table \ref{tlb:tta}. We can see that test-time collision handling decreases penetration significantly and saves a large number of grasps. 

\begin{table}
\centering
\caption{\textbf{Result of task-agnostic grasp.} $Emb_H$: hand embedding vector; T: test-time collision handling; pen: object penetration (cm); $\eta_f$: The collision-free grasp pose percentile.}
\label{tlb:tta}
\setlength{\tabcolsep}{2pt}
\begin{tabular}{ccccccc}
\toprule
\multirow{2}{*}{Method} & \multicolumn{3}{c}{seen obj}         & \multicolumn{3}{c}{unseen obj} \\ \cline{2-7} 
                        & $Q_1 \uparrow$ & pen$\downarrow$ & $\eta_f\uparrow$ & $Q_1 \uparrow$        & pen $\downarrow$ &   $\eta_f  \uparrow$   \\ \hline
DiffuTOG w. $Emb_H$           &   0.0533    &     0.578      &  6\%    &  \ralhigh{0.0416}      &      0.588    & 7\%       \\
\ralhigh{DiffuTOG w.o. $Emb_H$ + T }  &   \ralhigh{0.0664}    &     \ralhigh{0.872}     &  \ralhigh{39\%}    &  \ralhigh{0.0693}     &       \ralhigh{0.671}    & \ralhigh{39\%}      \\
DiffuTOG w. $Emb_H$ + T&   \textbf{0.1067}    &     \textbf{0.410}  &   \textbf{63\%}     &      \textbf{0.0930}        &  \textbf{0.385}  &  \textbf{50.5\%}   \\ \bottomrule 
\end{tabular}
\end{table}
\subsubsection{Diversity of Bootstrapping Data Augmentation}
Since we use a bootstrapping method to generate some challenging grasp poses, it's doubtful if the bootstrapping process just duplicates the existing pose instead of increasing the diversity of the data.
\ralhigh{To evaluate the diversity of grasp data, we normalized the joint angles and recorded the mean variance, mean range, and the number of valid grasps during each bootstrapping iteration. During the iteration, we observed that the mean variance initially increased from $5.6 \times 10^{-3}$ to $6.8 \times 10^{-3}$ and then slightly dropped to $6.4 \times 10^{-3}$. In addition, the mean range increases from $0.12$ to $0.42$, and the number of valid grasps increases from $3$ to $519$. By preserving the total number of generated grasps in each iteration, this bootstrapping method appears to enhance both the diversity and quantity of valid poses.}

\subsubsection{Success Rate of RL Policy} 
To demonstrate how the DexTOG loop improves the performance of RL policy, we compare the success rate of policy trained on the scratch dataset, the dataset only through a rule-based filter, and the dataset through the whole loop of DexTOG. The results are shown in Table \ref{tlb:loop_improve}.

\begin{table}
\centering
\setlength{\tabcolsep}{3pt}
\caption{Performance of RL policy on different data.}
\label{tlb:loop_improve}
\begin{tabular}{cccccc}
\toprule
Task & stapler & sprayer press & sprayer trigger & bottle & ballpoint\\ \midrule
Raw &  $59\%$ & $35\%$&  $5\%$ & $15\%$ & $10\%$\\
Rule-based  & $84\%$ & $50\%$ & $64\%$ & $79\%$ & $20\%$\\
Ours & $85\%$ & $57\%$ & $70\%$ & $81\%$ & $26\%$ \\ \bottomrule
\end{tabular}
\end{table}

\subsection{Limitation and Future Work}
Since DiffuTOG is a diffusion-based method, it is hard to alleviate the penetration by simply adding a penetration penalty to the loss function. The intermediate pose of the denoising process may be far from the final output pose, which may penetrate the object naturally. \ralhigh{Therefore, it's worth considering how to apply the commonly-used anti-penetration methods like the mentioned loss term and test-time augmentation.}

\ralhigh{
Besides, it's non-trival to make the state-based RL policy trained in simulation directly work in real-world settings, due to the sim-to-real gap.
Future works can focus on integrating some components like large vision-language models (VLMs) and domain randomization techniques into this framework to bridge the sim-to-real gap.
}

\section{Conclusion}
In this work, we propose a novel language-guided task-oriented dexterous grasp pose generation framework. The generated poses are evaluated by reinforcement learning algorithms in a physics-based simulator. The self-verification process inspires us to build a data engine that automatically generates task-oriented grasp poses for given objects and hand models. The quality of the generated task-oriented grasp poses is validated quantitatively and qualitatively. We hope the proposed method, data engine, and dataset can benefit task-oriented grasping research or more dynamic dexterous manipulation research.


%
%
\bibliographystyle{IEEEtran}
\bibliography{main}

\end{document}